\documentclass[10pt,twocolumn,letterpaper]{article}

\usepackage[pagenumbers]{cvpr} %

\usepackage{multirow}
\usepackage{colortbl}
\usepackage{lipsum}

\usepackage{algpseudocode}
\usepackage{bbm}
\usepackage[linesnumbered,ruled,vlined]{algorithm2e}
\usepackage{amsmath}
\definecolor{dark_green}{rgb}{0, 0.5, 0}

\definecolor{lightgray}{gray}{0.9}

\usepackage[dvipsnames]{xcolor}

\newcommand{\hlrow}{\rowcolor{black!6}}
\newcommand{\ours}{Seurat\xspace}
\newcommand{\paragrapht}[1]{\noindent\textbf{#1}}

\usepackage{pifont}

\definecolor{vividgreen}{RGB}{102,205,102}   %
\definecolor{vividred}{RGB}{255,99,71}       %
\definecolor{vividorange}{RGB}{255,165,79}   %

\usepackage{subcaption}
\usepackage{mathrsfs}

\newcommand{\myrowcolour}{}
\newcolumntype{a}{>{\myrowcolour}c}

\definecolor{cvprblue}{rgb}{0.21,0.49,0.74}
\usepackage[pagebackref,breaklinks,colorlinks,citecolor=cvprblue]{hyperref}

\def\paperID{12862} %
\def\confName{CVPR}
\def\confYear{2024}

\title{Seurat: From Moving Points to Depth}
\author{Seokju Cho\textsuperscript{1}\qquad
Jiahui Huang\textsuperscript{2}\qquad
Seungryong Kim\textsuperscript{1}\qquad
Joon-Young Lee\textsuperscript{2}\qquad\\[10pt]
\textsuperscript{1}KAIST AI\qquad \textsuperscript{2}Adobe Research\\[10pt]
\url{https://seurat-cvpr.github.io}
}

\begin{document}
\twocolumn[{%
\renewcommand\twocolumn[1][]{#1}%
\maketitle
\includegraphics[width=\linewidth]{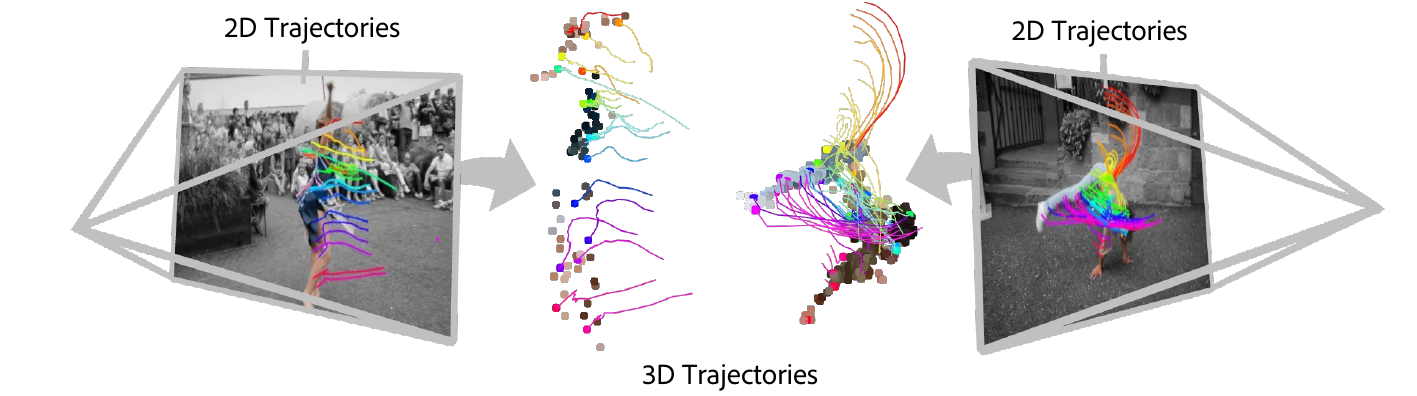}
\vspace{-20pt}
\captionof{figure}{\textbf{\ours} predicts precise and smooth depth changes for dynamic objects over time by only looking at the 2D point trajectories, which encode depth cues in their motion patterns. The figure illustrates 2D point tracks lifted into 3D space with our depth predictions on videos from the DAVIS dataset~\cite{pont20172017}.
}
\vspace{10pt}
\label{fig:teaser}
}]

\begin{abstract}
Accurate depth estimation from monocular videos remains challenging due to ambiguities inherent in single-view geometry, as crucial depth cues like stereopsis are absent. However, humans often perceive relative depth intuitively by observing variations in the size and spacing of objects as they move. Inspired by this, we propose a novel method that infers relative depth by examining the spatial relationships and temporal evolution of a set of tracked 2D trajectories. Specifically, we use off-the-shelf point tracking models to capture 2D trajectories. Then, our approach employs spatial and temporal transformers to process these trajectories and directly infer depth changes over time. Evaluated on the TAPVid-3D benchmark, our method demonstrates robust zero-shot performance, generalizing effectively from synthetic to real-world datasets. Results indicate that our approach achieves temporally smooth, high-accuracy depth predictions across diverse domains.
\end{abstract}

\begin{figure*}[t]
    \centering
    \begin{subfigure}[t]{0.53\linewidth}
        \centering
        \includegraphics[width=\linewidth]{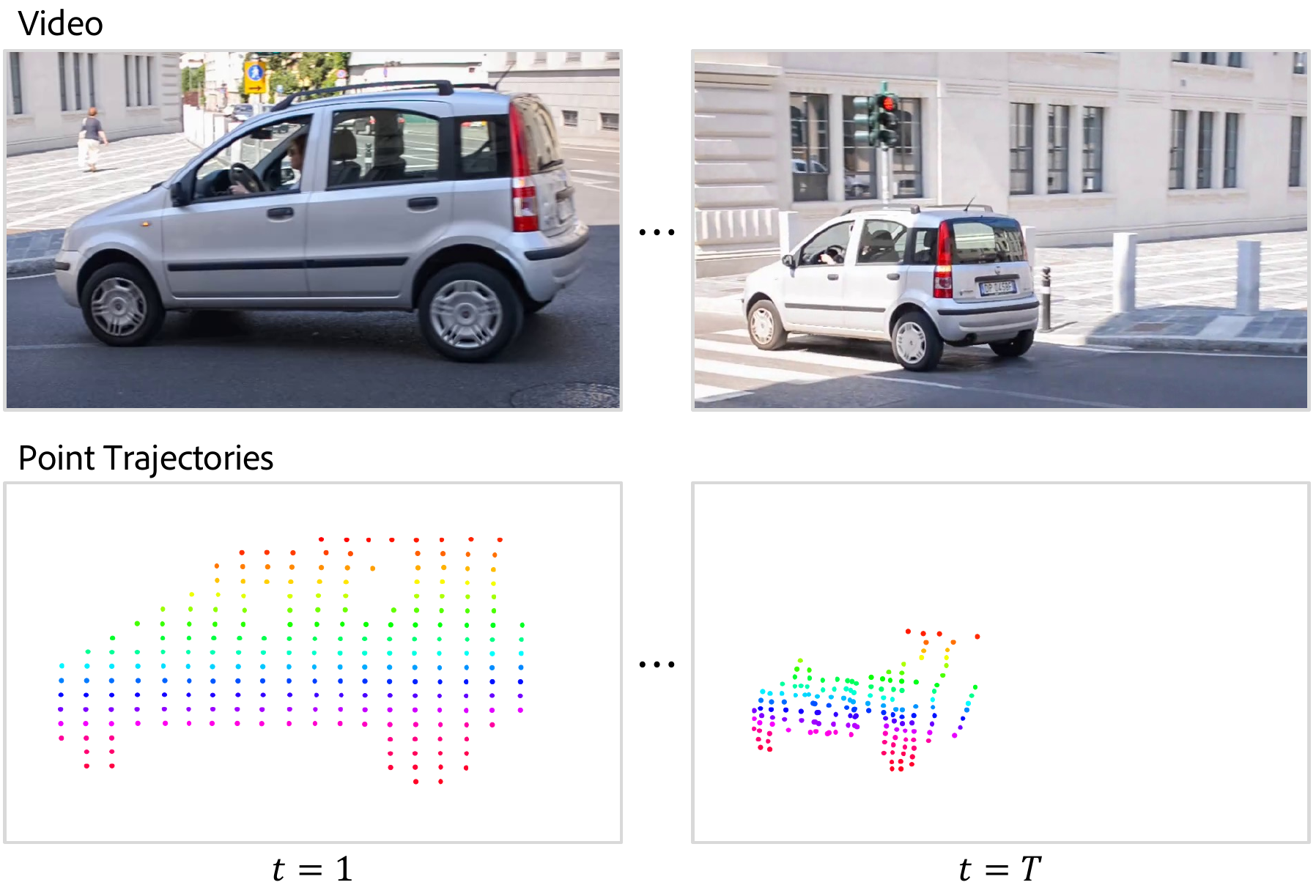}
        \caption{From the points alone, we can tell if the car is moving away or approaching.}
        \label{fig:motivation_1}
    \end{subfigure}
    \hfill
    \begin{subfigure}[t]{0.42\linewidth}
        \centering
        \includegraphics[width=\linewidth]{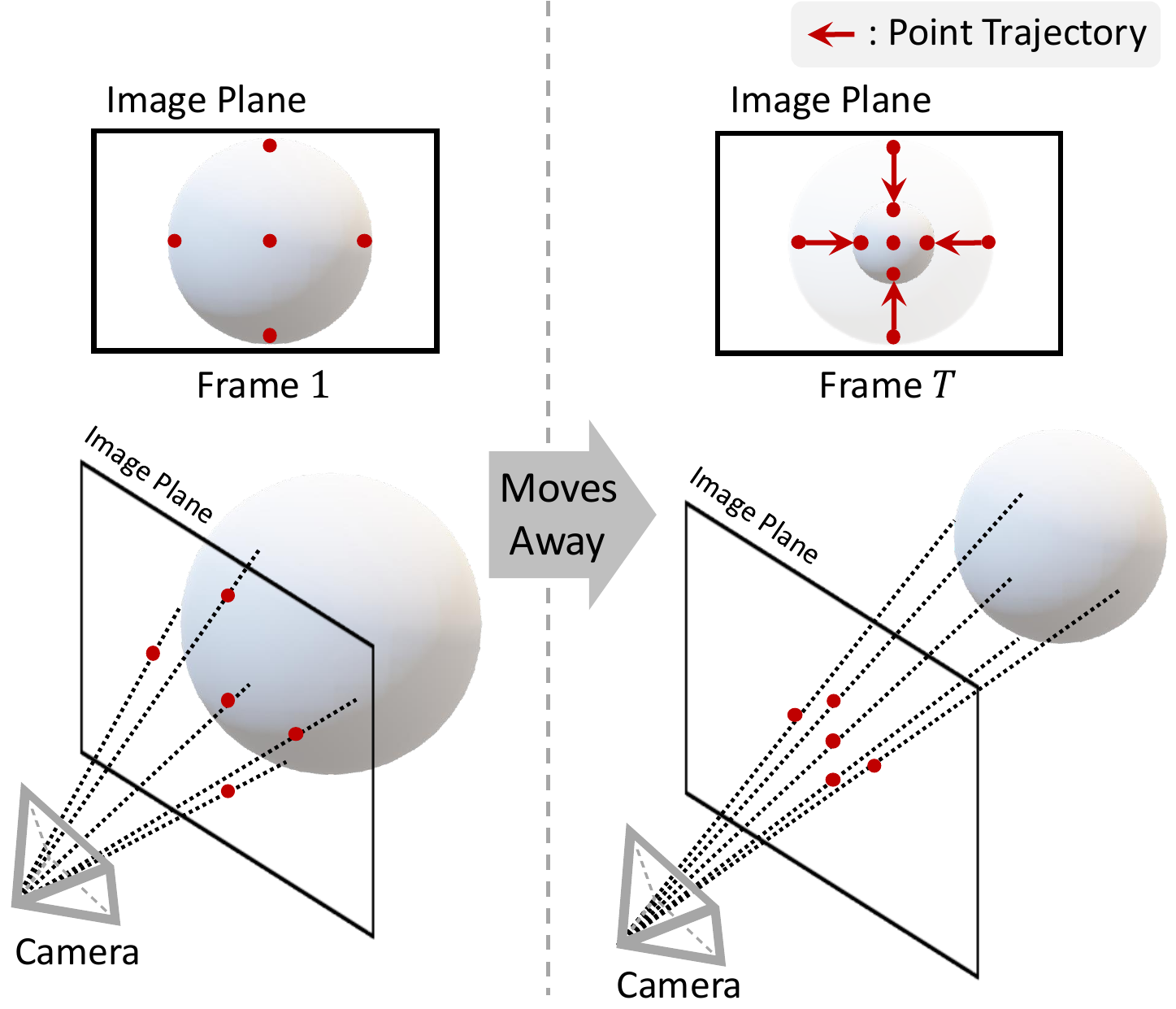}
        \caption{Object recedes, points converge toward center.}
        \label{fig:motivation_2}
    \end{subfigure}
    \vspace{-5pt}
    \caption{\textbf{Motivation of our work.} \textbf{(a)} By only looking at the tracked points, we can easily perceive that the object (here, a car) is moving away. \textbf{(b)} As a 3D object (here, a sphere) moves away from the camera, the pattern of its projected 2D points on the image plane changes, providing depth cues. In the initial frame (left), points are spaced farther apart on the image plane. As the object recedes (right), these 2D points converge toward the center, indicating increasing depth. This change in the density of projected points allows for inference of relative depth changes from motion in monocular video.}
    \label{fig:motivation}
    \vspace{-10pt}
\end{figure*}

\section{Introduction}

Understanding the 3D structure of a scene is essential for numerous applications, including image and video generation~\cite{zhang2023adding}, robotics~\cite{dong2022towards}, autonomous driving~\cite{pollefeys2008detailed}, and 3D reconstruction~\cite{li2023dynibar}. However, obtaining accurate depth information from monocular images is challenging due to inherent ambiguities~\cite{godard2017unsupervised}, difficulties in scale and shift determination~\cite{ranftl2020towards}, and the considerable diversity of real-world scenes~\cite{yang2024depth,ranftl2020towards}. In monocular video sequences, additional complexities arise from dynamic objects exhibiting intricate movements~\cite{luo2020consistent} and lengthy sequences complicating the maintenance of temporal coherence in depth estimation~\cite{hu2024depthcrafter}.

A classical method for obtaining precise depth information is Structured Light 3D Scanning~\cite{fechteler2007fast,fechteler2009adaptive,liu2010dual}, which involves projecting known patterns onto surfaces. The deformation of these patterns upon interacting with surfaces encodes valuable data about the 3D structure of the objects. This effectively transforms the spatial variations in the scene into measurable distortions in the projected pattern, allowing for accurate depth reconstruction.

Drawing inspiration from Structured Light methods~\cite{fechteler2007fast,fechteler2009adaptive,liu2010dual}, we propose that similar principles can be applied to monocular video sequences, assuming local object rigidity. Just as the deformation of a projected pattern reveals depth information, the patterns formed by the trajectories of tracked points~\cite{doersch2022tap,doersch2023tapir,karaev2023cotracker,cho2024local} in video sequences can reveal the 3D structure of a scene. These trajectories inherently capture spatial relationships and motion patterns relative to the camera, providing robust cues for depth estimation.
For example, points moving away from the camera create denser trajectory patterns, as illustrated in Figure~\ref{fig:motivation}. Analyzing the relative motion of these points allows us to discern whether objects or surfaces approach or recede from the camera, facilitating accurate temporal depth estimation.

Specifically, we begin by extracting trajectories using off-the-shelf point-tracking models~\cite{cho2024local,karaev2024cotracker}. To exploit depth information encoded in these trajectories, we employ both spatial and temporal transformers~\cite{arnab2021vivit,karaev2024cotracker}, which respectively model spatial relationships and ensure temporal smoothness. Furthermore, we explicitly decouple the supporting trajectories from query trajectories to avoid potential biases during depth estimation.

We observe that processing an entire video sequence simultaneously results in highly complex and unstable depth predictions, particularly for objects exhibiting rapid movements relative to the camera. To mitigate this issue, we predict temporal depth changes within sliding windows, assuming that depth variations are more consistent and manageable in shorter segments. Additionally, we introduce a specially designed window-wise log-ratio depth loss to achieve accurate depth supervision. We found this approach critical for reliably learning relative depth.

Our work introduces a novel trajectory-based framework for depth estimation from monocular videos, capitalizing on the temporal evolution of point trajectories without relying on stereo~\cite{chang2018pyramid} or multi-view setups~\cite{luiten2023dynamic3dgaussianstracking}, additional sensors~\cite{teed2021droid}, strong feature backbones~\cite{bochkovskii2024depth,oquab2023dinov2,leroy2024grounding}, or extensive annotated datasets~\cite{ranftl2020towards}. Our approach infers depth change over time in a strictly zero-shot manner, trained solely on a synthetic dataset~\cite{greff2022kubric} without any pre-trained feature backbone. Despite this simplicity, our model demonstrates robust generalization capabilities and performs effectively on real-world datasets.

We evaluate our proposed method on the TAPVid-3D benchmark~\cite{koppula2024tapvid}, highlighting its effectiveness and robustness in diverse scenarios, including driving scenes~\cite{Geiger2012CVPR,Geiger2013IJRR}, egocentric viewpoints~\cite{pan2023aria}, and videos containing deformations~\cite{joo2015panoptic}. Qualitative analyses further demonstrate that our method consistently produces temporally smooth and highly accurate depth predictions.

\section{Related Work}
\paragraph{Point tracking in 2D.} 
Track Any Point (TAP)~\cite{harley2022particle,koppula2024tapvid,doersch2023tapir,karaev2023cotracker,cho2024local,kim2025exploring}, or point tracking, aims to track any given query point throughout a video sequence along with its visibility status. PIPs~\cite{harley2022particle} iteratively refines trajectories using an MLP-Mixer~\cite{tolstikhin2021mlp} architecture. TAP-Net~\cite{doersch2022tap} constructs a global cost volume followed by convolutions and applies a soft-argmax operation for point tracking. TAPIR~\cite{doersch2023tapir} initializes trajectories with TAP-Net and refines them using PIPs' iterative refinement method. CoTracker~\cite{karaev2023cotracker} tracks multiple points simultaneously and models the interactions between them using a Transformer~\cite{vaswani2017attention} architecture. TAPTR~\cite{li2025taptr,li2024taptrv2} introduces the perspective of object detection to the point tracking. LocoTrack~\cite{cho2024local} enhances the cost volume with the neighboring points of query points, constructing 4D local correlation volume. Another approach involves test-time training~\cite{wang2023tracking,tumanyan2025dino,song2025track}, which can benefit from optimization regularization. In our work, we leverage these off-the-shelf point tracking models and extract depth information from the trajectories they produce.

\paragrapht{Point tracking in 3D.} Recently, SpatialTracker~\cite{xiao2024spatialtracker} demonstrated that point tracking can be improved by operating in 3D space. This is accomplished by lifting tracked points into 3D using a monocular depth estimation model~\cite{bhat2023zoedepth}, followed by iterative refinement within a triplane representation~\cite{chan2022efficient}, which enhances the accuracy of the resulting 2D point trajectories.
While SpatialTracker leverages 3D structural information to enable more robust tracking, our approach instead focuses on uncovering the 3D geometry inherently encoded within the trajectories themselves. TrackTo4D~\cite{kasten2024fast} performs Structure-from-Motion~\cite{schonberger2016structure} from trajectory data, representing dynamic parts with a low-dimensional basis~\cite{bregler2000recovering}. Unlike TrackTo4D, we do not introduce extra constraints or assumption for dynamic parts.

\paragrapht{Monocular depth estimation.}
With advancements in deep learning, MDE has achieved rapid progress by learning depth from large and diverse datasets. MegaDepth~\cite{li2018megadepth} demonstrated that training on extensive, varied datasets leads to better generalization and improved robustness to domain shifts. MiDaS~\cite{ranftl2020towards} built on this by blending datasets from multiple sources. Recent work has further expanded data sources by incorporating self-training on unlabeled data~\cite{yang2024depth} or using foundational models as a backbone, such as Stable Diffusion~\cite{rombach2022high} or DINOv2~\cite{oquab2023dinov2}. Although these approaches show strong results on per-frame depth estimation, they often show flickering when applied to video.

\section{Method}

\subsection{Motivation and Overview}

Estimating depth from monocular videos is challenging due to limited depth cues and inherent ambiguities in monocular images~\cite{godard2017unsupervised}. Traditional methods primarily focus on spatial relative depth~\cite{ranftl2020towards,bochkovskii2024depth,bhat2023zoedepth,yang2024depth}, which concerns the depth differences between points within the same frame.  
In contrast, we infer depth changes over time, exploiting the rich temporal information that monocular videos inherently contains.

We observe that point trajectories over time encapsulate valuable \emph{temporal depth} information. Specifically, the motion patterns of tracked points can reveal whether objects or surfaces are moving towards or away from the observer, as illustrated in Figure~\ref{fig:motivation}. Our approach leverages the temporal evolution of 2D point trajectories to predict depth changes, avoiding reliance on stereo~\cite{chang2018pyramid,teed2021droid}, multi-view setups~\cite{hartley2003multiple,leroy2024grounding}, additional sensors~\cite{teed2021droid}, or large-scale dataset~\cite{ranftl2020towards}.

Formally, given a monocular video with $T$ frames, we employ an off-the-shelf point tracking model~\cite{cho2024local,karaev2024cotracker} to extract \(N\) trajectories along with their occlusion statuses:

\begin{itemize} \item \textbf{Trajectories}: $\mathcal{T} = \{\mathbf{p}_i\}_{i=1}^N$, where each trajectory $\mathbf{p}_i = \{{p}_{i,t}\}_{t=1}^T$ consists of the 2D positions of point $i$ across frames. \item \textbf{Occlusion Statuses}: $\mathcal{V} = \{\mathbf{v}_i\}_{i=1}^N$, where $\mathbf{v}_i = \{{v}_{i,t}\}_{t=1}^T$ indicates the visibility of point $i$ at each frame. \end{itemize}

We develop a model that takes the extracted trajectories $\mathcal{T}$ and occlusion statuses $\mathcal{V}$ as input to predict temporal changes in depth. Specifically, our model estimates the depth ratio along each trajectory \(i\): $r_{i, t} = {d_{i, t}}/{d_{i, t_0}}$, relative to a reference frame $t_0$, where $d_{i, t}$ represents the depth of point $i$ at frame $t$.

We begin by detailing the theoretical basis behind extracting depth cues from trajectories in Sec.~\ref{sec:theory}. Subsequently, we introduce our Transformer-based model architecture for depth ratio prediction in Sec.~\ref{sec:transformer}. Next, we describe the sliding window training and inference in Sec.~\ref{sec:sliding-window} and outline the inference strategy in Sec.~\ref{sec:metric}, which combines predicted depth ratios $\hat{r}$ with a metric monocular depth estimation model to produce the final metric depth estimates $\hat{d}_{i, t}$.

\begin{figure}[t]
    \centering
    \includegraphics[width=1.0\linewidth]{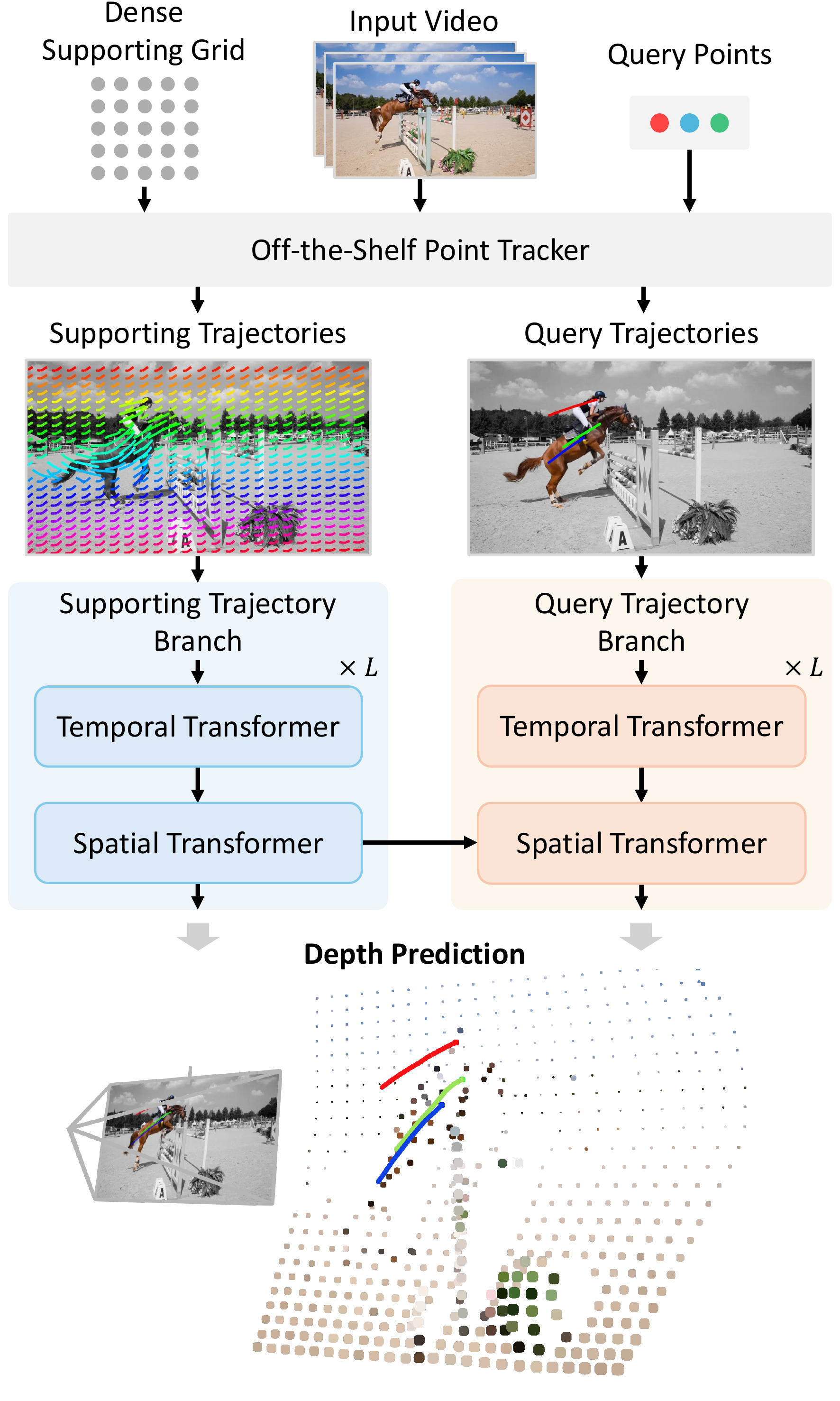}
    \vspace{-20pt}
    \caption{\textbf{Overall architecture.} We first use off-the-shelf point tracker~\cite{karaev2024cotracker,cho2024local} to extract 2D trajectories of query points and a dense supporting grid, then, these trajectories are processed with a temporal and a spatial transformer in two separate branches. Motion information encoded by the supporting branch is injected into the query branch via cross-attention. Finally, two regression heads output ratio depths of both supporting and query trajectories. }
    \label{fig:architecture}
    \vspace{-15pt}
\end{figure}

\subsection{Theoretical Analysis}\label{sec:theory}
Point trajectories over time encapsulate valuable temporal relative depth information. Specifically, the motion patterns of tracked points can reveal depth changes based on the \emph{density variation} of points projected onto the image plane.  Under the assumption of a pinhole camera model, consider a small surface patch with area \( A_{\text{surface}} \) at depth \( d_t \) and orientation \( \theta_t \) in frame \( t \), where \( \theta_t \) is the angle between the surface normal and the camera's optical axis. The projected area of this patch onto the image plane is given by:
\begin{equation}
A^{\text{image}}_t = \left( \frac{f}{d_t} \right)^2 A^{\text{surface}} \cos \theta_t,
\label{eq:projected_area}
\end{equation}
where \( f \) is the focal length of the camera. The density of projected points \( \rho^{\text{image}}_t \) is inversely proportional to \( A^{\text{image}}_t \):
\begin{equation}
\rho^{\text{image}}_t \propto \frac{(d_t)^2}{f^2 A_{\text{surface}} \cos \theta_t}.
\label{eq:density}
\end{equation}
Assuming local rigidity in a small area \( A^{\text{surface}} \), the ratio of densities at times \( t \) and \( t_0 \) is:
\begin{equation}
\frac{\rho^{\text{image}}_{t_0}}{\rho^{\text{image}}_t} = \left( \frac{d_{t_0}}{d_t} \right)^2 \left( \frac{\cos\theta_t}{\cos\theta_{t_0}} \right)^2.
\label{eq:density_ratio}
\end{equation}
Taking the square root yields the depth ratio \(r_t\), which is the depth variation over time with respect to the anchor timestep \(t_0\):
\begin{equation}
r_t = \frac{d_t}{d_{t_0}} = \sqrt{ \frac{\rho^{\text{image}}_t}{\rho^{\text{image}}_{t_0}} } \frac{\cos\theta_t}{\cos\theta_{t_0}}.
\label{eq:depth_ratio}
\end{equation}

It is important to note that the theoretical derivation above assumes the object is locally rigid and requires precise knowledge of its rotational orientation over time \( \cos\theta_t / \cos\theta_{t_0} \). Calculating depth changes deterministically using these formulas demands accurate measurements of rotation and local rigidity, which can be impractical in real-world scenarios, as validated in Table~\ref{tab:deterministic-baseline}. 

To overcome these limitations, we employ a transformer-based framework that implicitly captures the complexities of object motion and rotation through learned representations. This approach allows us to estimate ratio depth without relying solely on the explicit calculations provided by the equations, making the prediction more robust to variations in object properties and motion dynamics.

\subsection{Processing Trajectories with Transformer}\label{sec:transformer}
For effective depth estimation, it's crucial that the model captures the comprehensive motion patterns of the entire scene. Relying solely on user-defined or dataset-provided query points may not suffice, as they might not adequately represent the scene's overall motion due to their uneven distribution, often biased to the salient objects~\cite{doersch2022tap,zheng2023pointodyssey,balasingam2023drivetrackbenchmarklongrangepoint}. This limitation can hinder the model's ability to infer depth accurately, especially in regions lacking sufficient trajectory.

To address this challenge, we introduce \emph{supporting trajectories} derived from a grid of uniformly sampled points across the image. These grid-shaped trajectories offer a uniform and dense representation of the scene's motion, effectively capturing both local and global movements. By incorporating these supporting trajectories, the model gains a holistic understanding of the scene's dynamics, which is essential for accurate depth estimation.

Additionally, to prevent the biased distribution of query points from influencing the supporting trajectories, we decouple the model into two branches. The \textit{supporting trajectory branch} processes the supporting grid trajectories to capture global motion information, while the \textit{query trajectory branch} processes the query trajectories. The motion information encoded by the supporting branch is then injected into the query branch using cross-attention~\cite{vaswani2017attention}. This design ensures that the depth predictions for query points are informed by the overall scene motion without being biased by the distribution of the query points. The overall architecture is illustrated in Figure~\ref{fig:architecture}.

\subsubsection{Supporting Trajectory Branch}

The supporting trajectory branch processes a uniform grid of trajectories $\mathcal{T}_s$, which we refer to as \emph{supporting trajectories}. 
These are obtained by tracking a predefined grid of points using the point tracking model~\cite{cho2024local,karaev2024cotracker}. 
The supporting trajectories capture the overall motion dynamics of the scene and provide contextual information that aids in accurate depth estimation.

Let $\mathcal{T}_s = \{\mathbf{p}_{i}^{\text{supp}}\}_{i=1}^{i=N_s}$ denote the supporting trajectories, where $N_s$ is the number of supporting points. Each supporting trajectory \(\mathbf{p}_{i}^{\text{supp}}\) is associated with an occlusion status $\mathbf{v}_{i}^{\text{supp}}\in \mathcal{V}_s$. The encoder processes these trajectories using a Transformer architecture~\cite{vaswani2017attention} with alternating temporal and spatial attention layers~\cite{arnab2021vivit,karaev2024cotracker,Cho_2024_CVPR}.

Formally, the supporting trajectories are embedded and processed through $L$ layers:
\begin{align}
    \mathbf{h}^0_s &= \text{Embedding}(\mathcal{T}_s, \mathcal{V}_s)\nonumber\\
    \mathbf{h}_s^{l} &= \text{TransformerLayer}_s^{l}(\mathbf{h}_s^{l-1}), \quad l = 1, \dots, L,
\end{align}
where $\mathbf{h}_s^{0}$ is the initial embedding of the supporting trajectories from trajectory position and occlusion status. Each Transformer layer $\text{TransformerLayer}_s^{l}$ consists of temporal attention followed by spatial attention, following recent works in point tracking~\cite{karaev2024cotracker,Cho_2024_CVPR}. The intermediate features $\mathbf{h}_s^{l}$ captures the encoded motion information from the supporting trajectories, serving as a rich representation of the scene's dynamics.

\subsubsection{Query Trajectory Branch}

The query trajectory branch processes trajectories \(\mathcal{T}_q\) and their visibility status \(\mathcal{V}_q\), obtained by tracking user-defined query points. The query trajectory branch predicts the ratio depth by incorporating motion information from the supporting trajectories. To achieve this, we employ a Transformer architecture with cross-attention that attend to the motion information encoded from supporting trajectories. This allows the query decoder to leverage the global motion context while focusing on the specific query points.

The decoding process is defined as:
\begin{align}
    \mathbf{h}^0_q &= \mathrm{Embedding}(\mathcal{T}_q, \mathcal{V}_q)\nonumber\\
    \mathbf{h}_q^{l} &= \mathrm{TransformerLayer}_q^{l}(\mathbf{h}_q^{l-1}, \mathbf{h}_s^{l-1}), \quad l = 1, \dots, L,
\end{align}
where $\mathbf{h}_q^{0}$ is the initial embedding of the query trajectories derived from their positions and occlusion status. Each Transformer layer $\mathrm{TransformerLayer}_q^{l}$ includes a cross-attention that aggregates information from the supporting trajectories with \(\mathbf{h}_s^{l-1}\). The output $\mathbf{h}_q^{(L)}$ provides a refined representation for depth ratio prediction. Note that for each query point, trajectory is independently processed to prevent the predictions from being influenced by the distribution of query points.

Both the supporting encoder and the query decoder have ratio depth prediction heads attached to their final layers, which output the estimated relative depth changes for the trajectories.
Additionally, inspired by~\cite{teed2020raft, harley2022particle, Cho_2024_CVPR, cho2024local,karaev2024cotracker}, we iteratively refine the predicted depths by feeding the current predictions back into the model in subsequent iterations. 

\subsection{Sliding Window Prediction}\label{sec:sliding-window}

Processing entire video sequences at once may lead to complex and divergent depth results due to the increased complexity of long-range motion patterns and the potential for supporting trajectories to move out of the frame. To address this, we employ a \emph{sliding window} approach during both training and inference.

The sliding window approach involves segmenting the video sequence into shorter temporal windows of length $W$, with an overlap to ensure continuity, setting the window stride as \(S\).
In our model, the query trajectories persist across windows, while the supporting trajectories are re-initialized for each window, making the supporting trajectories more likely to remain within the frame.

\paragrapht{Training with window-wise log ratio depth loss.}
To align our training objective with Eq.~\ref{eq:depth_ratio}, we focus on predicting the \emph{log depth ratio} for each frame with respect to the starting frame of the current window. 
By predicting the log depth ratios, the model becomes invariant to the absolute scale of depth, focusing instead on the relative changes.
The log depth ratio for point \( i \) at time \( t \) within window \( w \) is defined as:
\begin{equation} 
    \ell^w_{i, t} = \log \left( r_{i,t}^w \right) = \log \left( \frac{d^w_{i,t}}{d^w_{i,0}} \right), \quad t \in [0, W - 1],
\end{equation}
where \( W \) is the window size, and \( d_{i,t}^w \) is the ground-truth depth of point \( i \) at time~\( t \) within \(w\)-th window.

Our model predicts \(\hat{\ell}_{i,t}^w\), the predicted log depth ratios for each point \( i \) at time \( t \) within window \( w \). We train the model using an \( L_1 \) loss between the predicted and ground-truth log depth ratios \(\ell_{i,t}\):
\begin{equation}
    \mathcal{L} = \sum_{i} \sum_w \sum_t \left| \hat{\ell}_{i,t}^w - \ell_{i,t}^w \right|.
\end{equation}
This loss encourages the model to accurately predict the depth changes within each window, focusing on the temporal relative depth. We apply the loss to the depth predictions of both the \emph{query trajectories} and the \emph{supporting trajectories}. 

\paragrapht{Inference with sliding window.}
Within each sliding window, the model predicts log depth ratios, denoted as $\hat{\ell}_{i,t}^w$, relative to the first frame of that window. To obtain depth predictions for the entire sequence, these log depth ratios are accumulated, and then exponentiated to convert them back to linear-scale depth ratios. Importantly, during inference, the supporting point depth ratios are discarded after each window; only the query point depth ratios are accumulated, as the supporting points serve solely to improve depth estimation within the current window. Specifically, the accumulated depth ratio $\hat{r}_{i,t}$ for point $i$ at time $t$ is calculated as:
\begin{equation}
\hat{r}_{i,t} = \exp\left( \hat{\ell}_{i,t_k}^{k} + \sum_{w=0}^{k-1} \hat{\ell}_{i, S}^{w} \right), \quad t_k = t-Sk,
\end{equation}
where $\hat{\ell}_{i,t}^{w}$ represents the predicted log depth ratio for \(t\)-th frame in window $w$, $S$ denotes the stride of the sliding window, $k$ is the index of the last window that contains time $t$, and $t_k$ represents the time index of \(t\) within window $k$.

\begin{table*}[t]
    \centering
    \resizebox{\textwidth}{!}{%
    \setlength{\tabcolsep}{3pt}
    \begin{tabular}{l|c|ccc|ccc|ccc|ccc}
    \toprule
    Depth & \multirow{2}{*}{Point Tracker}
          & \multicolumn{3}{c|}{Aria~\cite{pan2023aria}}
          & \multicolumn{3}{c|}{DriveTrack~\cite{balasingam2023drivetrackbenchmarklongrangepoint}}
          & \multicolumn{3}{c|}{PStudio~\cite{joo2015panoptic}}
          & \multicolumn{3}{c}{\textbf{Average}} \\
    Estimator &
          & 3D-AJ $\uparrow$ & APD $\uparrow$ & TC $\downarrow$
          & 3D-AJ $\uparrow$ & APD $\uparrow$ & TC $\downarrow$
          & 3D-AJ $\uparrow$ & APD $\uparrow$ & TC $\downarrow$
          & 3D-AJ $\uparrow$ & APD $\uparrow$ & TC $\downarrow$ \\
    \midrule\midrule

    \textcolor{gray}{Oracle depth}$^*$ & CoTracker~\cite{karaev2024cotracker}
      & \textcolor{gray}{55.9} & \textcolor{gray}{70.3} & -
      & \textcolor{gray}{53.2} & \textcolor{gray}{71.7} & -
      & \textcolor{gray}{46.9} & \textcolor{gray}{65.0} & -
      & \textcolor{gray}{52.0}    & \textcolor{gray}{69.0}    & - \\
    \midrule

    TAPIR-3D~\cite{koppula2024tapvid} & --
      & 8.5  & 14.9 & -
      & 10.2 & 17.0 & -
      & 7.2  & 13.1 & -
      & 8.6  & 15.0 & - \\
    \midrule

    \multicolumn{14}{c}{\textit{Per-frame Depth Estimator}}\\

    \multirow{3}{*}{ZoeDepth~\cite{bhat2023zoedepth}}
    & \textcolor{gray}{Oracle tracker}$^*$
      & \textcolor{gray}{19.1} & \textcolor{gray}{28.6} & \textcolor{gray}{0.05}
      & \textcolor{gray}{10.8} & \textcolor{gray}{17.4} & \textcolor{gray}{1.27}
      & \textcolor{gray}{15.5} & \textcolor{gray}{24.3} & \textcolor{gray}{0.05}
      & \textcolor{gray}{15.1} & \textcolor{gray}{23.4} & \textcolor{gray}{0.46} \\

    & CoTracker~\cite{karaev2024cotracker}
      & 16.5 & 25.5 & 0.06
      & 9.5  & 16.2 & 1.34
      & 11.8 & 20.4 & 0.05
      & 12.6 & 20.7 & 0.48 \\

    & LocoTrack~\cite{cho2024local}
      & 15.7 & 25.4 & 0.06
      & 10.0 & 16.3 & 1.33
      & 11.8 & 20.1 & 0.04
      & 12.5 & 20.6 & 0.48 \\

    \midrule
    \multirow{3}{*}{DepthPro~\cite{bochkovskii2024depth}}
    & \textcolor{gray}{Oracle tracker}$^*$
      & \textcolor{gray}{13.4} & \textcolor{gray}{21.3} & \textcolor{gray}{0.14}
      & \textcolor{gray}{6.7}  & \textcolor{gray}{12.0} & \textcolor{gray}{3.72}
      & \textcolor{gray}{8.9}  & \textcolor{gray}{15.3} & \textcolor{gray}{0.16}
      & \textcolor{gray}{9.7}  & \textcolor{gray}{16.2} & \textcolor{gray}{1.34} \\

    & CoTracker~\cite{karaev2024cotracker}
      & 11.3 & 18.7 & 0.16
      & 5.4  & 10.5 & 3.87
      & 6.7  & 12.7 & 0.16
      & 7.8  & 14.0 & 1.40 \\

    & LocoTrack~\cite{cho2024local}
      & 10.8 & 18.6 & 0.15
      & 5.6  & 10.3 & 4.18
      & 6.7  & 12.5 & 0.17
      & 7.7  & 13.8 & 1.50 \\

    \midrule

    \multicolumn{14}{c}{\textit{Video Depth Estimator}}\\

    \multirow{3}{*}{DepthCrafter~\cite{hu2024depthcrafter}}
    & \textcolor{gray}{Oracle tracker}$^*$
      & \textcolor{gray}{17.5} & \textcolor{gray}{27.1} & \textcolor{gray}{0.02}
      & \textcolor{gray}{9.8}  & \textcolor{gray}{16.4} & \textcolor{gray}{0.75}
      & \textcolor{gray}{13.8} & \textcolor{gray}{22.6} & \textcolor{gray}{0.03}
      & \textcolor{gray}{13.7} & \textcolor{gray}{22.0} & \textcolor{gray}{0.26} \\

    & CoTracker~\cite{karaev2024cotracker}
      & 15.1 & 24.3 & 0.04
      & 8.4  & 15.2 & 0.98
      & 11.1 & 19.5 & \underline{0.03}
      & 11.5 & 19.7 & 0.35 \\

    & LocoTrack~\cite{cho2024local}
      & 13.9 & 23.5 & \underline{0.03}
      & 8.8  & 15.1 & 0.95
      & 11.0 & 19.3 & \underline{0.03}
      & 11.2 & 19.3 & 0.34 \\

    \midrule
    \multirow{3}{*}{ChronoDepth~\cite{shao2024learning}}
    & \textcolor{gray}{Oracle tracker}$^*$
      & \textcolor{gray}{12.3} & \textcolor{gray}{18.7} & \textcolor{gray}{0.22}
      & \textcolor{gray}{7.1}  & \textcolor{gray}{12.7} & \textcolor{gray}{3.94}
      & \textcolor{gray}{3.5}  & \textcolor{gray}{6.5}  & \textcolor{gray}{0.33}
      & \textcolor{gray}{7.6}  & \textcolor{gray}{12.6} & \textcolor{gray}{1.50} \\

    & CoTracker~\cite{karaev2024cotracker}
      & 11.0 & 17.8 & 2.37
      & 6.1  & 11.7 & 7.43
      & 3.0  & 5.9  & 0.37
      & 6.7  & 11.8 & 3.39 \\

    & LocoTrack~\cite{cho2024local}
      & 10.5 & 17.6 & 0.23
      & 6.4  & 11.6 & 1.49
      & 3.0  & 5.9  & 0.38
      & 6.6  & 11.7 & 0.70 \\

    \midrule
    \hlrow
    & CoTracker~\cite{karaev2024cotracker}
      & \textbf{25.1} & \textbf{36.9} & \textbf{0.01}
      & \underline{11.6} & \textbf{20.4} & \textbf{0.15}
      & \underline{17.3} & \textbf{28.5} & \textbf{0.01}
      & \underline{18.0} & \textbf{28.6} & \textbf{0.05} \\

    \hlrow
    \multirow{-2}{*}{\ours (Ours)}
    & LocoTrack~\cite{cho2024local}
      & \underline{24.1} & \underline{34.6} & 0.08
      & \textbf{12.7} & \textbf{20.4} & \underline{0.55}
      & \textbf{17.4} & \underline{27.6} & 0.07
      & \textbf{18.1} & \underline{27.4} & \underline{0.23} \\

    \bottomrule
    \end{tabular}
    }
    \vspace{-7pt}
    \caption{\textbf{Quantitative results on TAPVid-3D~\cite{koppula2024tapvid} minival split with per-trajectory depth scaling.} 
    \textcolor{gray}{\emph{Oracle tracker}}$^*$ rows use ground-truth 2D trajectories, while \textcolor{gray}{\emph{Oracle depth}}$^*$ row uses ground-truth depth estimation to determine the upper bound.}
    \label{table:per_trajectory_baselines}
    \vspace{-15pt}
\end{table*}

\subsection{Incorporating Depth Ratio and Metric Depth}\label{sec:metric}
Although our predicted depth ratios accurately capture changes over time, the accumulated ratio \(\hat{r}_{i, t}\) for the entire video reflects only the depth change relative to the initial frame. Therefore, in the final step, we correct each accumulated depth ratio using an off-the-shelf metric depth estimator~\cite{bhat2023zoedepth,bochkovskii2024depth}.

Concretely, we perform \emph{piecewise scale matching} for each visible subsequence \(\mathcal{S}_{i,t}\). Here, \(\mathcal{S}_{i,t}\) is the set of consecutive frames that includes time \(t\), during which a particular point \(i\) remains visible. We compute a scale factor \(s_{i,t}\) by matching the median of \(\hat{r}_{i, t}\) to the median of the MDE model’s depth estimates over all \(t' \in \mathcal{S}_{i,t}\):
\begin{equation}
s_{i,t} = \frac{ 
  \mathrm{median}_{\,t' \,\in\, \mathcal{S}_{i,t}}\!\Bigl(d_{\text{MDE}}\bigl(p_{i,t'}\bigr)\Bigr) 
}{
  \mathrm{median}_{\,t' \,\in\, \mathcal{S}_{i,t}}\!\bigl(\hat{r}_{i,t'}\bigr)
},
\end{equation}
where \(d_{\text{MDE}}(p_{i,t'})\) is the MDE-predicted depth at point \(p_{i,t'}\), obtained by bilinear interpolation from the depth map. We then apply this scale factor to our predicted ratio to obtain the final depth estimate:
\begin{equation}
\hat{d}_{i,t} = s_{i,t} \,\cdot\, \hat{r}_{i,t}.
\end{equation}

By matching medians in each subsequence, we align our model’s temporal depth changes with the MDE model’s metric depth estimates. This integration leverages the strengths of both approaches: our model offers temporally coherent depth changes, while the MDE model ensures reliable spatial depth relationships within each frame. The resulting estimates are therefore both temporally stable and spatially precise.

\begin{table*}[t]
    \centering
    \resizebox{\textwidth}{!}{%
    \setlength{\tabcolsep}{3pt}
    \begin{tabular}{l|c|ccc|ccc|ccc|ccc}
    \toprule
    Depth & \multirow{2}{*}{Point Tracker}
          & \multicolumn{3}{c|}{Aria~\cite{pan2023aria}}
          & \multicolumn{3}{c|}{DriveTrack~\cite{balasingam2023drivetrackbenchmarklongrangepoint}}
          & \multicolumn{3}{c|}{PStudio~\cite{joo2015panoptic}}
          & \multicolumn{3}{c}{\textbf{Average}} \\
    Estimator &
          & 3D-AJ $\uparrow$ & APD $\uparrow$ & TC $\downarrow$
          & 3D-AJ $\uparrow$ & APD $\uparrow$ & TC $\downarrow$
          & 3D-AJ $\uparrow$ & APD $\uparrow$ & TC $\downarrow$
          & 3D-AJ $\uparrow$ & APD $\uparrow$ & TC $\downarrow$ \\
    \midrule\midrule
    
    \textcolor{gray}{Oracle depth}$^*$ & CoTracker~\cite{karaev2024cotracker}
      & \textcolor{gray}{55.9} & \textcolor{gray}{70.3} & -
      & \textcolor{gray}{53.2} & \textcolor{gray}{71.7} & -
      & \textcolor{gray}{46.9} & \textcolor{gray}{65.0} & -
      & \textcolor{gray}{52.0}    & \textcolor{gray}{69.0}    & - \\
    \midrule
    
    \multirow{3}{*}{DepthCrafter~\cite{hu2024depthcrafter}}
    & \textcolor{gray}{Oracle tracker}$^*$
      & \textcolor{gray}{9.5}  & \textcolor{gray}{15.1} & \textcolor{gray}{0.016}
      & \textcolor{gray}{9.6}  & \textcolor{gray}{15.2} & \textcolor{gray}{0.484}
      & \textcolor{gray}{13.4} & \textcolor{gray}{21.1} & \textcolor{gray}{0.015}
      & \textcolor{gray}{10.8} & \textcolor{gray}{17.1} & \textcolor{gray}{0.172} \\

    & CoTracker~\cite{karaev2024cotracker}
      & 7.4  & 12.2 & 0.041
      & 7.2  & 12.3 & 0.594
      & 9.3  & 16.4 & 0.015
      & 8.0  & 13.6 & 0.217 \\

    & LocoTrack~\cite{cho2024local}
      & 6.7  & 12.2 & \underline{0.030}
      & 7.0  & 11.5 & 0.584
      & 8.2  & 14.5 & 0.017
      & 7.3  & 12.7 & 0.210 \\
    \midrule
    \multirow{3}{*}{ChronoDepth~\cite{shao2024learning}}
    & \textcolor{gray}{Oracle tracker}$^*$
      & \textcolor{gray}{12.3} & \textcolor{gray}{18.7} & \textcolor{gray}{0.033}
      & \textcolor{gray}{8.0}  & \textcolor{gray}{13.3} & \textcolor{gray}{0.913}
      & \textcolor{gray}{11.8} & \textcolor{gray}{18.9} & \textcolor{gray}{0.020}
      & \textcolor{gray}{10.7} & \textcolor{gray}{17.0} & \textcolor{gray}{0.322} \\

    & CoTracker~\cite{karaev2024cotracker}
      & 10.1 & 15.9 & 0.045
      & 6.7  & 11.9 & 0.895
      & 8.6  & 15.4 & 0.020
      & 8.5  & 14.4 & 0.320 \\

    & LocoTrack~\cite{cho2024local}
      & 9.4  & 15.7 & 0.044
      & 6.2  & 10.7 & 0.888
      & 8.3  & 14.7 & 0.022
      & 8.0  & 13.7 & 0.318 \\
    \midrule
    \hlrow
    {\ours (Ours)} & CoTracker~\cite{karaev2024cotracker}
      & 11.3 & 18.0 & \textbf{0.012}
      & 7.5  & 12.9 & \underline{0.244}
      & 11.4 & \underline{19.2} & \textbf{0.012}
      & 10.1 & 16.7 & \underline{0.089} \\

    \hlrow
    \quad+ ZoeDepth~\cite{bhat2023zoedepth}
    & LocoTrack~\cite{cho2024local}
      & 10.7 & 16.7 & 0.086
      & 7.6  & 12.3 & 0.567
      & 10.4 & 16.8 & 0.069
      & 9.6  & 15.3 & 0.241 \\
    \midrule
    \hlrow
    {\ours (Ours)} & CoTracker~\cite{karaev2024cotracker}
      & \textbf{15.1} & \textbf{22.5} & \textbf{0.012}
      & \underline{8.4} & \textbf{13.9} & \textbf{0.219}
      & \textbf{12.5} & \textbf{20.5} & \underline{0.013}
      & \textbf{12.0} & \textbf{19.0} & \textbf{0.081} \\

    \hlrow
    \quad+ DepthPro~\cite{bochkovskii2024depth}
    & LocoTrack~\cite{cho2024local}
      & \underline{14.5} & \underline{21.4} & 0.086
      & \textbf{8.7} & \underline{13.5} & 0.551
      & \underline{12.0} & 18.8 & 0.070
      & \underline{11.7} & \underline{17.9} & 0.236 \\
    \bottomrule
    \end{tabular}%
    }

    \vspace{-7pt}
    \caption{\textbf{Quantitative results of affine-invariant depth on TAPVid-3D~\cite{koppula2024tapvid} minival split.} Compared to video depth estimators, our model shows superior performance. 
    \textcolor{gray}{\emph{Oracle tracker}}$^*$ rows use ground-truth 2D trajectories, while \textcolor{gray}{\emph{Oracle depth}}$^*$ row uses ground-truth depth estimation to determine the upper bound.}
    \label{table:scale-and-shift}
    \vspace{-15pt}
\end{table*}

\begin{table}[t]
    \centering
    
    \begin{subtable}[t]{0.625\linewidth}
        \centering
        \resizebox{\linewidth}{!}{
            \begin{tabular}{ll|cc}
                \toprule
                & \multirow{2}{*}{Method} & \multicolumn{2}{c}{Average} \\
                & & 3D-AJ & APD \\
                \midrule\midrule
                \textbf{(I)} & \ours{} (Ours) & \textbf{18.0} & \textbf{28.6} \\
                \textbf{(II)} & \textbf{(I)} - Two-branch design & {13.7} & {23.1} \\
                \textbf{(III)} & \textbf{(I)} - Sliding window & {8.8} & {16.1} \\
                \textbf{(IV)} & \textbf{(I)} - Window-wise loss & \underline{14.9} & \underline{24.5} \\
                \bottomrule
            \end{tabular}
        }
        \label{tab:main-ablation}
    \end{subtable}
    \hfill
    \begin{subtable}[t]{0.355\linewidth}
        \centering
        \resizebox{\linewidth}{!}{
            \begin{tabular}{l|cc}
                \toprule
                \# of & \multicolumn{2}{c}{Average} \\
                Layers & 3D-AJ & APD \\
                \midrule\midrule
                1 & {15.5} & {25.1} \\
                2 (Ours) & \textbf{18.0} & \textbf{28.6} \\
                3 & \underline{17.8} & \underline{28.3} \\
                4 & \underline{17.8} & {28.1} \\
                \bottomrule
            \end{tabular}
        }
        \label{tab:ablation-layer}
    \end{subtable}
    \vspace{-7pt}
    \caption{\textbf{Ablation studies.} Left: Ablation of main components. We conduct gradually exclude core components from our full model. Right: Ablation on the number of layers $L$.}
    \label{tab:combined-ablation}
    \vspace{-10pt}
\end{table}

\begin{table}[t]
\centering
\resizebox{\linewidth}{!}{%
\begin{tabular}{l|c|cc|cc|cc|cc}
\toprule
Depth Ratio & \multirow{2}{*}{Point Tracker} 
& \multicolumn{2}{c|}{Aria} 
& \multicolumn{2}{c|}{DriveTrack} 
& \multicolumn{2}{c|}{PStudio} 
& \multicolumn{2}{c}{\textbf{Average}} \\
Estimator && 3D-AJ $\uparrow$ & APD $\uparrow$
          & 3D-AJ $\uparrow$ & APD $\uparrow$
          & 3D-AJ $\uparrow$ & APD $\uparrow$
          & 3D-AJ $\uparrow$ & APD $\uparrow$ \\
\midrule\midrule
Hand-crafted baseline & CoTracker
  & \underline{8.6} & \underline{15.2}
  & \underline{5.3} & \underline{10.4}
  & \underline{4.2} & \underline{8.1}
  & \underline{6.0} & \underline{11.2} \\
\midrule
\hlrow \ours (Ours) & CoTracker
  & \textbf{25.1} & \textbf{36.9}
  & \textbf{11.6} & \textbf{20.4}
  & \textbf{17.3} & \textbf{28.5}
  & \textbf{18.0} & \textbf{28.6} \\
\bottomrule
\end{tabular}
}
\vspace{-7pt}
\caption{\textbf{Comparison with handcrafted baseline.} Handcrafted implementation of Eq.~\ref{eq:depth_ratio} exhibits lower performance.}
\label{tab:deterministic-baseline}
\vspace{-15pt}
\end{table}

\section{Experiment}
\subsection{Evaluation Protocol and Baselines}
\paragraph{Evaluation protocol.} We assess trajectory depth accuracy using the TAPVid-3D benchmark~\cite{koppula2024tapvid}, which encompasses both outdoor and indoor scenarios through egocentric videos, driving scenes, and studio setups. For details on the datasets included in the TAPVid-3D benchmark, please refer to the supplementary materials.

We measure the accuracy of the predicted position using \textbf{APD}, which represents the percentage of points within a threshold \(\delta\). Unlike 2D point tracking, which defines threshold values in pixels, we use a depth-adaptive threshold~\cite{koppula2024tapvid}. We report the average score across \(\delta=1,2,4,8,16\). 
\textbf{3D-AJ} reflects combined accuracy for both position and occlusion. In this study, we prioritize position accuracy over occlusion, as we rely on the occlusion accuracy of an off-the-shelf 2D point tracker. We also measure the temporal coherence (\textbf{TC})~\cite{wang2023tracking,Cho_2024_CVPR} of the predicted 3D tracks, which is the \(L2\) distance between the ground-truth acceleration of the trajectory and that of the predicted trajectory.

\paragrapht{Baselines.} For comparison, we construct simple baselines by pairing a 2D point tracking model with a monocular depth estimation (MDE) model. By unprojecting the 2D tracks generated by the tracking module using depth estimates from the MDE model, we obtain a 3D trajectory over the video sequence. For 2D point tracking, we use recent state-of-the-art models for their accuracy, specifically LocoTrack~\cite{cho2024local} and CoTracker~\cite{karaev2024cotracker}. We carefully select MDE models, ensuring that their training sets do not overlap with our evaluation benchmark dataset. 
Some state-of-the-art MDE models, such as UniDepth~\cite{piccinelli2024unidepth}, use the Waymo dataset~\cite{Sun_2020_CVPR} for training. To avoid overlap, we use ZoeDepth~\cite{bhat2023zoedepth} and DepthPro~\cite{bochkovskii2024depth} as our baseline models.

\subsection{Implementation Details}
We train our model using the AdamW optimizer~\cite{loshchilov2017decoupled} with a learning rate of $5 \times 10^{-4}$ and a weight decay of $1 \times 10^{-5}$. We linearly decayed the learning rate during training, with warm-up step of 1,000. We conduct training for 100,000 steps on eight NVIDIA RTX 3090 GPUs, using a batch size of 1 per GPU. We generate 90,000 training samples from the Kubric~\cite{greff2022kubric} MOVi-F dataset generator. We use the number of Transformer layers~\cite{vaswani2017attention} as \(L=2\), where each layer has a hidden dimension of 384 and uses 8 attention heads. The supporting grid size is set to $24 \times 24$. For LocoTrack~\cite{cho2024local}, we use the base model with a resolution of \(256\times 256\). For CoTracker~\cite{karaev2024cotracker}, we use the global grid of \(6 \times 6\). We set the temporal window size \(W=8\).

\subsection{Main Results}
\paragraph{Quantitative results.}
We conduct a quantitative evaluation on the TAPVid-3D benchmark in Table~\ref{table:per_trajectory_baselines} and Table~\ref{table:scale-and-shift}. For comparison, we combine existing point tracking methods~\cite{cho2024local, karaev2024cotracker} with monocular depth estimators~\cite{bhat2023zoedepth, bochkovskii2024depth} and video depth estimators~\cite{hu2024depthcrafter,shao2024learning} by unprojecting the tracked 2D trajectories into 3D space using the predicted depth maps. Despite the simplicity of this approach, their strong tracking and depth performance provide a robust baseline. 

\begin{table}[t]
    \centering
    \resizebox{\linewidth}{!}{%
    \setlength{\tabcolsep}{3pt}
    \begin{tabular}{l|c|cc|cc|cc|cc}
    \toprule
    \multirow{2}{*}{Depth Ratio Estimator} & \multirow{2}{*}{Point Tracker} 
    & \multicolumn{2}{c|}{Aria} 
    & \multicolumn{2}{c|}{DriveTrack} 
    & \multicolumn{2}{c|}{PStudio} 
    & \multicolumn{2}{c}{\textbf{Average}} \\

    & & 3D-AJ $\uparrow$ & APD $\uparrow$ 
      & 3D-AJ $\uparrow$ & APD $\uparrow$ 
      & 3D-AJ $\uparrow$ & APD $\uparrow$ 
      & 3D-AJ $\uparrow$ & APD $\uparrow$ \\
    \midrule\midrule

    \ours + Texture patch 
    & CoTracker 
    & \underline{21.0} & \underline{31.8}
    & \underline{10.3} & \underline{18.6}
    & \underline{16.1} & \underline{27.1}  
    & \underline{15.8} & \underline{25.8}  \\
    \midrule
    \hlrow
    \ours (Ours)
    & CoTracker 
    & \textbf{25.1} & \textbf{36.9} 
    & \textbf{11.6} & \textbf{20.4} 
    & \textbf{17.3} & \textbf{28.5} 
    & \textbf{18.0} & \textbf{28.6}  \\
    \bottomrule
    \end{tabular}
    }
    \vspace{-7pt}
    \caption{\textbf{Texture patch ablation.} Additional texture information reduces performance.}
    \label{tab:texture-patch}
    \vspace{-10pt}
\end{table}

\begin{table}[t]
    \centering
    \resizebox{\linewidth}{!}{%
    \begin{tabular}{l|cc|cc|cc}
        \toprule
         \multirow{2}{*}{Method} & \multicolumn{2}{c|}{Aria} & \multicolumn{2}{c|}{DriveTrack} & \multicolumn{2}{c}{PStudio} \\
           & 3D-AJ $\uparrow$ & APD $\uparrow$ & 3D-AJ $\uparrow$ & APD $\uparrow$ & 3D-AJ $\uparrow$ & APD $\uparrow$ \\
        \midrule\midrule
         ZoeDepth + CoTracker & $\underline{9.8}$ & $\underline{15.8}$ & $\underline{7.2}$ & $\underline{12.3}$ & $\underline{10.2}$ & $\underline{17.9}$ \\
        ZoeDepth + CoTracker + 1 iter. of Gaussian smoothing  & $4.8$ & $8.6$ & $6.3$ & $11.1$ & $7.7$ & $14.1$ \\
        ZoeDepth + CoTracker + 3 iters. of Gaussian smoothing  & $4.4$ & $8.0$ & $6.0$ & $10.7$ & $7.4$ & $13.6$ \\
\midrule
\hlrow\ours (Ours) & ${\textbf{14.6}}$ & ${\textbf{21.9}}$& $\textbf{6.9}$ & $\textbf{11.8}$ &  $\textbf{12.7}$ & $\textbf{20.7}$  \\
        \bottomrule
    \end{tabular}%
    }
    \vspace{-7pt}
    \caption{\textbf{Comparison to simple Gaussian smoothing.} Simple Gaussian smoothing of per-frame depth estimates proves insufficient for achieving precise video depth.}
    \label{tab:smoothing}
    \vspace{-15pt}
\end{table}

\begin{figure*}[t]
    \centering
    
    \includegraphics[width=1.0\linewidth]{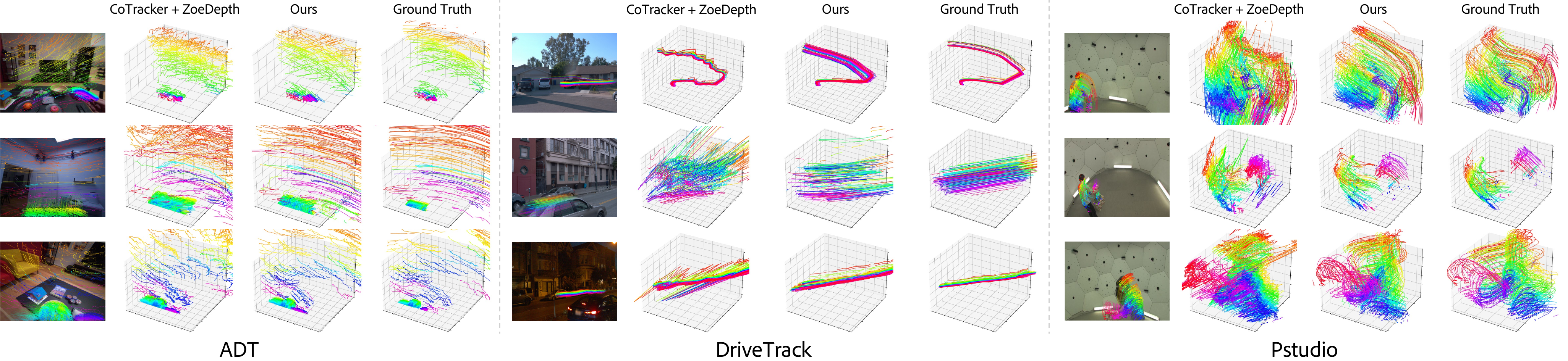}
    \vspace{-20pt}
    \caption{\textbf{Qualitative comparisons to baselines.} We visualize 3D trajectories using the TAPVid-3D~\cite{koppula2024tapvid} benchmark. Compared to baselines that use combinations such as CoTracker with ZoeDepth, our model achieves superior depth smoothness and accuracy.}
    \label{fig:qualitative}
    \vspace{-15pt}
\end{figure*}

Table~\ref{table:per_trajectory_baselines} presents the tracking precision evaluated on a per-trajectory basis, with each trajectory scaled according to the depth of its query point. By individually scaling each trajectory, we can assess how accurately our method captures depth changes over time for each point, aligning with the focus of our approach. Our method significantly outperforms other baselines in position precision (APD), especially in the Aria dataset. Notably, unlike other depth prediction models~\cite{bhat2023zoedepth,bochkovskii2024depth} trained on 14-21 large-scale annotated real datasets, our method \textit{trained on single synthetic dataset demonstrates more robust generalizability} in depth prediction. Additionally, it excels in temporal coherence, particularly on the DriveTrack dataset, achieving more than \(40\times\) better coherence, underscoring the stability of our approach.

Table~\ref{table:scale-and-shift} presents the tracking precision obtained when predicted depths are adjusted to ground truth depths using a single scale and shift value per video, determined via least squares. This evaluation is similar to the common practice in video depth estimation studies~\cite{hu2024depthcrafter,shao2024learning}. This metric reflects the accuracy of depth predictions across an entire video sequence. Our method achieves strong results on both the Aria and PStudio datasets, particularly when combined with DepthPro~\cite{bochkovskii2024depth}. Additionally, we show that our model provides even better temporal coherence compared to existing video depth estimation methods.

\paragrapht{Qualitative results.} In Figure~\ref{fig:qualitative}, we compare our method with baseline models. Our model demonstrates exceptionally smooth depth predictions with high accuracy relative to the ground truth.

\subsection{Ablation and Analysis}
\paragraph{Component ablation.}
We conduct an ablation study, presented on the left of Table~\ref{tab:combined-ablation}. In \textbf{(II)}, we remove the two-branch design (query and supporting trajectory branches) of our architecture, instead processing both supporting and query trajectories together using a Transformer encoder. In \textbf{(III)}, we eliminate the sliding window approach, processing the entire video sequence at once. In \textbf{(IV)}, rather than using ratio depth with respect to the first frame of the sliding window, we use ratio depth with respect to the query point.

The observed performance degradation when omitting each design choice demonstrates the effectiveness of our approach. Specifically, in \textbf{(II)}, joint processing of supporting and query points negatively impacts performance. In \textbf{(III)}, processing the full sequence at once significantly reduces performance, suggesting that handling the whole sequence may complicate the learning process due to complex motion patterns. Finally, \textbf{(IV)} shows that predicting ratio depth within a sliding window is beneficial for the model.

\paragrapht{Ablation on the number of layers \(L\).} On the left of Table~\ref{tab:combined-ablation}, we show the score with varying number of Transformer layers. We found that more than 2 layers does not guarantees the performance boost. 

\paragrapht{Analysis on hand-crafted deterministic baseline.} In Table~\ref{tab:deterministic-baseline}, we validate the handcrafted baseline that implements Eq.~\ref{eq:depth_ratio}. To measure how the local spatial density of semi-densely sampled trajectories evolves over time, we use a kernel-based approach that compares each point's neighborhood structure across frames. Specifically, for each trajectory point, we fix its \(k\)-nearest neighbors based on their positions in the first frame and then compute a kernel density estimate in each frame using Gaussian kernels applied to the Euclidean distances between the point and its fixed neighbors in that frame. This provides a smooth, robust estimate of local density that reflects how tightly clustered a point remains with respect to its original neighborhood over time. The density at each frame is then normalized by the corresponding density in the first frame, yielding a relative density change per point across the temporal window.

The results imply that the hand-crafted baseline performs significantly worse than ours, which we believe is due to its sensitivity to prediction noise and its inability to estimate surface normals, demonstrating the impracticality of estimating trajectory density using a handcrafted method.

\paragrapht{Analysis of the impact of texture as input.} To investigate whether texture information can help the model infer depth changes, we conduct an analysis by adding an RGB input path to our model, as shown in Table~\ref{tab:texture-patch}. Specifically, we extract local RGB patches around the trajectory, flatten them, and concatenate them with the Transformer input. The results show that incorporating texture information actually degrades performance. We suspect this is because training on texture data may lead to overfitting to the synthetic dataset.

\paragrapht{Is \ours simply smoothing the per-frame depth jitter?} Table~\ref{tab:smoothing} addresses whether \ours is just trajectory smoothing. Simple Gaussian smoothing of the lifted trajectory, using per-frame depth estimates, yielded worse results than \ours, indicating that \ours enforces temporal geometric consistency beyond simple smoothing.

\section{Conclusion}
We introduce a novel trajectory-based framework for depth estimation that leverages the temporal evolution of point trajectories in monocular videos. Using spatial and temporal transformers within a sliding window approach, our method processes trajectory data to accurately capture depth changes with high temporal smoothness. This approach offers a fresh perspective on 3D understanding from monocular videos—simple, effective, and straightforward to implement. We hope this method will inspire further exploration and innovation in related fields.

\twocolumn[{%
\renewcommand\twocolumn[1][]{#1}%
\maketitlesupplementary
\resizebox{\textwidth}{!}{%
\begin{tabular}{l|c|ccc|ccc|ccc|ccc}
\toprule
Depth & \multirow{2}{*}{Point Tracker} 
      & \multicolumn{3}{c|}{Aria~\cite{pan2023aria}} 
      & \multicolumn{3}{c|}{DriveTrack~\cite{balasingam2023drivetrackbenchmarklongrangepoint}} 
      & \multicolumn{3}{c|}{PStudio~\cite{joo2015panoptic}} 
      & \multicolumn{3}{c}{\textbf{Average}} \\
Estimator & 
      & 3D-AJ $\uparrow$ & APD $\uparrow$ & TC $\downarrow$
      & 3D-AJ $\uparrow$ & APD $\uparrow$ & TC $\downarrow$ 
      & 3D-AJ $\uparrow$ & APD $\uparrow$ & TC $\downarrow$ 
      & 3D-AJ $\uparrow$ & APD $\uparrow$ & TC $\downarrow$ \\
\midrule\midrule

\textcolor{gray}{Oracle depth}$^*$ & CoTracker~\cite{karaev2024cotracker}
    & $\textcolor{gray}{55.9}$ & $\textcolor{gray}{70.3}$ & - 
    & $\textcolor{gray}{53.2}$ & $\textcolor{gray}{71.7}$ & - 
    & $\textcolor{gray}{46.9}$ & $\textcolor{gray}{65.0}$ & - 
    & $\textcolor{gray}{52.0}$ & $\textcolor{gray}{69.0}$ & - \\
\midrule

\multirow{5}{*}{ZoeDepth~\cite{bhat2023zoedepth}} 

& \textcolor{gray}{Oracle tracker}$^*$  
  & \textcolor{gray}{11.8} & \textcolor{gray}{18.6} & \textcolor{gray}{0.04}
  & \textcolor{gray}{8.6} & \textcolor{gray}{13.9} & \textcolor{gray}{1.22}
  & \textcolor{gray}{14.2} & \textcolor{gray}{22.2} & \textcolor{gray}{0.05}
  & \textcolor{gray}{11.5} & \textcolor{gray}{18.2} & \textcolor{gray}{0.44} \\

& CoTracker~\cite{karaev2024cotracker}  
  & 9.8 & 15.8 & 0.06
  & \underline{7.2} & \textbf{12.3} & \underline{1.26}
  & 10.2 & 17.9 & \underline{0.05}
  & 9.1 & 15.3 & \underline{0.46} \\

& LocoTrack~\cite{cho2024local} 
  & 9.6 & 15.7 & \underline{0.05}
  & \textbf{7.5} & \textbf{12.3} & 1.33
  & 9.9 & 17.3 & 0.06
  & 9.0 & 15.1 & 0.48 \\

& SpatialTracker~\cite{xiao2024spatialtracker}  
  & 9.2 & 15.1 & -
  & 5.8 & 10.2 & -
  & 9.8 & 17.7 & -
  & 8.3 & 14.3 & - \\

\hlrow + \ours (Ours) & CoTracker
  & \underline{10.9} & \underline{18.8} & \textbf{0.01}
  & 6.6 & 11.6 & \textbf{0.27}
  & \underline{10.9} & \underline{18.8} & \textbf{0.01}
  & \underline{9.6} & \underline{16.4} & \textbf{0.10} \\
  
\midrule
\multirow{4}{*}{DepthPro~\cite{bochkovskii2024depth}}

& \textcolor{gray}{Oracle tracker}$^*$ 
  & \textcolor{gray}{12.2} & \textcolor{gray}{18.9} & \textcolor{gray}{0.13}
  & \textcolor{gray}{7.0} & \textcolor{gray}{12.0} & \textcolor{gray}{3.72}
  & \textcolor{gray}{11.0} & \textcolor{gray}{18.0} & \textcolor{gray}{0.16}
  & \textcolor{gray}{10.1} & \textcolor{gray}{16.3} & \textcolor{gray}{1.34} \\

& CoTracker~\cite{karaev2024cotracker} 
  & 9.9 & 15.9 & 0.15
  & 5.2 & 9.4 & 3.38
  & 7.8 & 14.4 & 0.16
  & 7.6 & 13.2 & 1.23 \\

& LocoTrack~\cite{cho2024local} 
  & 9.2 & 15.6 & 0.13
  & 5.3 & 9.3 & 3.60
  & 7.8 & 14.2 & 0.17
  & 7.4 & 13.0 & 1.30 \\

\hlrow + \ours (Ours) & CoTracker
  & \textbf{14.6} & \textbf{21.9} & \textbf{0.01}
  & 6.9 & \underline{11.8} & \textbf{0.27}
  & \textbf{12.7} & \textbf{20.7} & \textbf{0.01}
  & \textbf{11.4} & \textbf{18.1} & \textbf{0.10} \\
\bottomrule
\end{tabular}}

\vspace{-5pt}
\captionof{table}{\textbf{Quantitative results on TAPVid-3D~\cite{koppula2024tapvid} minival split with meidan scaling.} We combined the depth ratio from \ours with the metric depth predictions from ZoeDepth~\cite{bhat2023zoedepth} and DepthPro~\cite{bochkovskii2024depth}. \textcolor{gray}{\emph{Oracle tracker}}$^*$ rows use ground-truth 2D trajectories, while \textcolor{gray}{\emph{Oracle depth}}$^*$ row uses ground-truth depth estimation to determine the upper bound.
}\label{table:median-scaling}
\vspace{5pt}
}]

The supplementary materials begin with an additional quantitative comparison in Sec.~\ref{sec:video-depth}.
Next, we present an analysis of inference time in Sec.~\ref{sec:inference-time}. Sequentially, Sec.~\ref{sec:more-details} provides additional implementation details. Finally, we discuss the limitations of our work in Sec.~\ref{sec:limitation}.

\section{More Results}
\paragraph{Quantitative comparison with median scaling.}\label{sec:video-depth}
In Table~\ref{table:median-scaling}, we compare our method with baselines that combine depth estimators~\cite{bhat2023zoedepth,bochkovskii2024depth} and point trackers~\cite{karaev2023cotracker,cho2024local}. Overall, our method outperforms the baselines, with particularly significant improvements over those using DepthPro. Additionally, our method demonstrates substantially better temporal coherency (TC), further highlighting its effectiveness in maintaining consistent depth predictions over time.

\begin{table*}[t]
    \centering
    \setlength{\tabcolsep}{3pt}
    \begin{tabular}{l|c|cc|cc|cc|cc}
    \toprule
    Depth & \multirow{2}{*}{Point Tracker} & \multicolumn{2}{c|}{Aria~\cite{pan2023aria}} & \multicolumn{2}{c|}{DriveTrack~\cite{balasingam2023drivetrackbenchmarklongrangepoint}} & \multicolumn{2}{c|}{PStudio~\cite{joo2015panoptic}} & \multicolumn{2}{c}{\textbf{Average}} \\
    Estimator & & AbsRel $\downarrow$ & $\delta_1$ $\uparrow$ & AbsRel $\downarrow$ & $\delta_1$ $\uparrow$ & AbsRel $\downarrow$ & $\delta_1$ $\uparrow$ & AbsRel $\downarrow$ & $\delta_1$ $\uparrow$ \\
    \midrule\midrule
    \multirow{3}{*}{DepthCrafter~\cite{hu2024depthcrafter}} & \textcolor{gray}{Oracle tracker}$^*$ & {\color{gray}$0.344$} & {\color{gray}$0.564$} & {\color{gray}$0.141$} & {\color{gray}$0.811$} & {\color{gray}$0.053$} & {\color{gray}$0.981$} & {\color{gray}$0.179$} & {\color{gray}$0.785$} \\
    & CoTracker~\cite{karaev2024cotracker} & $0.390$ & $0.542$ & $0.165$ & $0.787$ & $0.057$ & $0.974$ & $0.204$ & $0.768$ \\
    & LocoTrack~\cite{cho2024local} & $0.394$ & $0.529$ & $0.182$ & $0.784$ & $0.058$ & $0.973$ & $0.211$ & $0.762$ \\
    \midrule
    \multirow{3}{*}{ChronoDepth~\cite{shao2024learning}} & \textcolor{gray}{Oracle tracker}$^*$ & {\color{gray}$0.248$} & {\color{gray}$0.671$} & {\color{gray}$0.106$} & {\color{gray}$0.887$} & {\color{gray}$0.064$} & {\color{gray}$0.977$} & {\color{gray}$0.139$} & {\color{gray}$0.845$} \\
    & CoTracker~\cite{karaev2024cotracker} & $0.287$ & $0.644$ & $0.132$ & $0.843$ & $0.067$ & $0.971$ & $0.162$ & $0.819$ \\
    & LocoTrack~\cite{cho2024local} & $0.290$ & $0.648$ & $0.166$ & $0.811$ & $0.068$ & $0.971$ & $0.175$ & $0.810$ \\
    \midrule
    \hlrow {\ours (Ours)} & CoTracker~\cite{karaev2024cotracker} & $\textbf{0.198}$ & $\textbf{0.754}$ & $\textbf{0.127}$ & $\textbf{0.868}$ & $\textbf{0.052}$ & $\textbf{0.984}$ & $\textbf{0.126}$ & $\textbf{0.869}$ \\
    \hlrow \quad+ ZoeDepth~\cite{bhat2023zoedepth} & LocoTrack~\cite{cho2024local} & $0.223$ & $0.732$ & $0.161$ & $0.838$ & $0.056$ & $0.981$ & $0.147$ & $0.850$ \\
    \bottomrule
    \end{tabular}%
    \vspace{-5pt}
    \caption{\textbf{Quantitative results of affine-invariant depth on TAPVid-3D~\cite{koppula2024tapvid} minival split.} \textcolor{gray}{\emph{Oracle tracker}}$^*$ rows use ground-truth 2D trajectories to determine the upper bound}
    \label{table:depth-metric-video}
    \vspace{-5pt}
\end{table*}

\begin{table*}[t]
    \centering
    \setlength{\tabcolsep}{3pt}
    \begin{tabular}{l|c|cc|cc|cc|cc}
    \toprule
    Depth & \multirow{2}{*}{Point Tracker} & \multicolumn{2}{c|}{Aria~\cite{pan2023aria}} & \multicolumn{2}{c|}{DriveTrack~\cite{balasingam2023drivetrackbenchmarklongrangepoint}} & \multicolumn{2}{c|}{PStudio~\cite{joo2015panoptic}} & \multicolumn{2}{c}{\textbf{Average}} \\
    Estimator & & AbsRel $\downarrow$ & $\delta_1$ $\uparrow$ & AbsRel $\downarrow$ & $\delta_1$ $\uparrow$ & AbsRel $\downarrow$ & $\delta_1$ $\uparrow$ & AbsRel $\downarrow$ & $\delta_1$ $\uparrow$ \\
    \midrule\midrule
    \multirow{3}{*}{ZoeDepth~\cite{bhat2023zoedepth}} & \textcolor{gray}{Oracle tracker}$^*$ & {\color{gray}$0.252$} & {\color{gray}$0.698$} & {\color{gray}$0.122$} & {\color{gray}$0.863$} & {\color{gray}$0.057$} & {\color{gray}$0.973$} & {\color{gray}${0.144}$} & {\color{gray}${0.845}$} \\
    & CoTracker~\cite{karaev2024cotracker} & $0.277$ & $0.671$ & $0.149$ & $0.817$ & $0.062$ & $0.967$ & $0.163$ & $0.818$ \\
    & LocoTrack~\cite{cho2024local} & $0.282$ & $0.676$ & $0.175$ & $0.799$ & $0.063$ & $0.966$ & $0.173$ & $0.814$ \\
    \midrule
    \hlrow {\ours (Ours)} & CoTracker~\cite{karaev2024cotracker} & $\textbf{0.244}$ & $\textbf{0.701}$ & $\textbf{0.139}$ & $\textbf{0.845}$ & $\textbf{0.060}$ & $\textbf{0.967}$ & $\textbf{0.148}$ & $\textbf{0.838}$\\
    \hlrow \quad+ ZoeDepth~\cite{bhat2023zoedepth} & LocoTrack~\cite{cho2024local} & $0.266$ & $0.694$ & $0.178$ & $0.818$ & $0.063$ & $0.964$ & $0.169$ & $0.825$ \\
    \midrule\midrule
    \multirow{3}{*}{DepthPro~\cite{bhat2023zoedepth}} & \textcolor{gray}{Oracle tracker}$^*$ & {\color{gray}$0.180$} & {\color{gray}$0.742$} & {\color{gray}$0.148$} & {\color{gray}$0.827$} & {\color{gray}$0.067$} & {\color{gray}$0.970$} & {\color{gray}${0.132}$} & {\color{gray}${0.846}$} \\
    & CoTracker~\cite{karaev2024cotracker} & $0.219$ & $0.708$ & $0.186$ & $0.778$ & $0.074$ & $0.958$ & $0.160$ & $0.815$ \\
    & LocoTrack~\cite{cho2024local} & $0.214$ & $0.720$ & $0.214$ & $0.760$ & $0.076$ & $0.955$ & $0.168$ & $0.812$ \\
    \midrule
    \hlrow {\ours (Ours)} & CoTracker~\cite{karaev2024cotracker} & $\textbf{0.179}$ & $0.757$ & $\textbf{0.153}$ & $\textbf{0.833}$ & $\textbf{0.055}$ & $\textbf{0.976}$ & $\textbf{0.129}$ & $\textbf{0.855}$\\
    \hlrow \quad+ DepthPro~\cite{bochkovskii2024depth} & LocoTrack~\cite{cho2024local} & $0.198$ & $\textbf{0.760}$ & $0.182$ & $0.810$ & $0.060$ & $0.967$ & $0.147$ & $0.846$ \\
    \bottomrule
    \end{tabular}%
    
    \vspace{-5pt}
    \caption{\textbf{Quantitative results on TAPVid-3D~\cite{koppula2024tapvid} using depth metrics with median scaling.} \textcolor{gray}{\emph{Oracle tracker}}$^*$ rows use ground-truth 2D trajectories to determine the upper bound.}
    \label{table:depth-metric-image}
    \vspace{-10pt}
\end{table*}

\begin{table}[t]
    \centering
    \resizebox{\linewidth}{!}{
   \begin{tabular}{c|l|ccccc}
        \toprule
         &Method& 1 point  & 10 points & $10^2$ points & $10^3$ points & $10^4$ points
         \\
        \midrule\midrule
        \textbf{(I)} &Tracking  & 2.07& 2.02& 2.02& 4.58& 34.98\\

        \textbf{(II)} &Depth Inference  & 0.24& 0.24& 0.24& 0.47& 4.28\\
        \midrule
        \hlrow\textbf{(III)} & Overall  & 2.31& 2.26& 2.26& 5.05& 39.26 \\
        \bottomrule
\end{tabular}}%
\vspace{-5pt}
        \caption{\textbf{Inference time and the number of query points.} We measure how the inference time (s) for a 24 frame video changes as varying the number of query points. We measure the time for point tracking (\textbf{I}) and depth inference with our method (\textbf{II}) separately. We use CoTracker~\cite{karaev2024cotracker} as a point tracker. Inference time is measured using Nvidia RTX 3090 GPU.}
    \label{tab:analysis-runtime}
    \vspace{-10pt}
\end{table}

\paragrapht{Quantitative comparison using depth metrics.}\label{sec:depth-metric}
In Table~\ref{table:depth-metric-video} and Table~\ref{table:depth-metric-image}, we compare our method with other baselines using metrics widely adopted in depth estimation literature~\cite{ranftl2020towards,hu2024depthcrafter,shao2024learning,yang2024depth}. Specifically, we employ the absolute relative error (AbsRel) and \(\delta_1\). AbsRel is calculated as \(|\hat{d} - d|/d\), while \(\delta_1\) is defined as the percentage of \(\text{max}\left(\hat{d}/d, d/\hat{d}\right) < 1.25\), where \(\hat{d}\) denotes the predicted depth and \(d\) denotes the ground-truth depth. In Table~\ref{table:depth-metric-video}, since the compared depth estimators~\cite{hu2024depthcrafter,shao2024learning} predict affine-invariant depth, we apply scale-and-shift optimization using least squares to align their predicted depth scales with the ground truth. For \ours that uses the trajectory of CoTracker~\cite{karaev2024cotracker} as input consistently outperforms other baselines significantly, validating its effectiveness in depth accuracy. 

\section{More Analysis}

\paragraph{Analysis on inference time.}\label{sec:inference-time}
Table~\ref{tab:analysis-runtime} presents an analysis of inference time with different numbers of query points. We separately measure the inference time required for point tracking and depth inference using our method. The results show that point tracking accounts for most of the computation time, while our model is relatively efficient. We believe that future advancements in point tracking efficiency will lead to more efficient inference for our overall pipeline.

\section{More Implementation Details}\label{sec:more-details}
During training, we sample \(N_q=256\) query trajectories per batch. While we utilize trajectories from off-the-shelf models~\cite{karaev2024cotracker,cho2024local} during inference, we use ground-truth trajectory positions and occlusion information as input during training. Occluded positions in the input trajectories are masked by replacing their values with the last visible position.  In addition to the depth prediction head at the end of our model, we include a head to predict the position of occluded points, using the same loss function as in~\cite{karaev2024cotracker}. We found this beneficial for the model to produce smooth depth estimates for occluded points. For iterative depth refinement, we consistently use 4 iterations for both training and inference.

\section{Limitations and Discussion}\label{sec:limitation}
We have shown that the temporal evolution of depth can be inferred from trajectories extracted by off-the-shelf point trackers. However, our model has a limited ability to infer spatial relative depth, relying instead on the monocular depth estimation model. End-to-end training of this combined pipeline could further synergize temporal and spatial depth estimation in video, which we leave as future research. 
Furthermore, our use of a sliding window approach for processing long video sequences, while making depth variation manageable, somewhat limits the potential benefits of longer sequences. Exploring alternative approaches to effectively leverage extended temporal information is another interesting area for future research.

\paragrapht{Acknowledgements.} This research was supported by Institute of Information \& communications Technology Planning \& Evaluation (IITP) grant funded by the Korea government (MSIT) (RS-2019-II190075, RS-2024-00509279, RS-2025-II212068, RS-2023-00227592) and the Culture, Sports, and Tourism R\&D Program through the Korea Creative Content Agency grant funded by the Ministry of Culture, Sports and Tourism (RS-2024-00345025, RS-2024-00333068), and National Research Foundation of Korea (RS-2024-00346597).

\clearpage

{
    \small
    \bibliographystyle{ieeenat_fullname}
    \bibliography{main}
}

\end{document}